\documentclass[preprint,12pt]{elsarticle}

\usepackage{amssymb}
\usepackage{booktabs}
\usepackage{xcolor}  
\usepackage{amsmath}

\newcommand{\rev}[1]{{\color{black} #1}}

\journal{Neurocomputing}

\begin{document}

\begin{frontmatter}



\title{Iterated Relevance Matrix Analysis (IRMA) for \rev{the} identification of class-discriminative subspaces}


\author[inst1,inst2]{Sofie L{\"o}vdal\corref{cor1}}
\ead{s.s.lovdal@rug.nl}
\cortext[cor1]{Corresponding author}

\affiliation[inst1]{organization={Bernoulli Institute for Mathematics, Computer Science and Artificial Intelligence},
            addressline={Nijenborgh 9}, 
            city={Groningen},
            postcode={9747AG}, 
            country={Netherlands}}

\author[inst1,inst3]{Michael Biehl}
\ead{m.biehl@rug.nl}

\affiliation[inst2]{organization={University Medical Center Groningen, Department of Nuclear Medicine and Molecular Imaging},
            addressline={Hanzeplein 1}, 
            city={Groningen},
            postcode={9713GZ}, 
            country={Netherlands}}

\affiliation[inst3]{organization={SMQB, Institute of Metabolism and Systems Research, College of Medical and Dental Sciences, University of Birmingham},
addressline={Address}, 
city={Birmingham},
postcode={B152TT}, 
country={United Kingdom}}

\begin{abstract}
\rev{We introduce and investigate the iterated application of Generalized Matrix Learning Vector Quantizaton} for the analysis of feature relevances in classification problems, \rev{as well as for the construction of class-discriminative subspaces.} The suggested Iterated Relevance Matrix Analysis (IRMA) identifies
a linear subspace representing the classification specific information of the considered data sets using Generalized Matrix Learning Vector Quantization \rev{(GMLVQ)}. By iteratively determining a new discriminative subspace while projecting out all previously identified ones, a combined subspace carrying all class-specific information can be found. This facilitates a detailed analysis of feature relevances, and \rev{enables} improved low-dimensional representations and visualizations of labeled data sets. Additionally, the IRMA-based class-discriminative subspace can be used for dimensionality reduction and
\rev{the training of robust classifiers with potentially improved performance.} 
\end{abstract}


\begin{highlights}
\item IRMA identifies a linear subspace representing the class-specific information in \rev{a given} data set by iteratively retraining GMLVQ classifiers
\item \rev{Improves interpretation of feature relevances in classification problems, enabling the identification and  comparison of  multiple possible solutions}
\item We train and apply simple GLVQ classifiers in original feature space, GMLVQ-space and IRMA-space to illustrate the potential of using IRMA for dimensionality reduction
\item Exploiting the resulting class-discriminative subspace, as well as each individual model resulting from an IRMA iteration, are both promising for the purpose of improving performance of classifiers
\end{highlights}


\end{frontmatter}


\section{Introduction}
Prototype-based systems such as Learning Vector Quantization \rev{(LVQ) \cite{kohonen1986learning, kohonen1995learning, sato1995generalized, nova2014review} can } serve 
as genuinely interpretable and transparent classification tools
\cite{ghosh2020visualisation}. In combination
with the use of adaptive distance measures \cite{schneider2009adaptive,bunte2012limited},
they provide valuable insights into the
structure of the problem at hand and into the relevance of features for the actual classification task. However, 
the presence of correlated features or multiple \rev{subsets of features enabling similar performance can lead to ambiguous relevance assignments
and non-unique outcomes of training}. This 
frequently complicates the interpretation of relevance learning, see e.g. \cite{gopfert2017feature,gopfert2018interpretation}. Similarly, a classifier trained by gradient descent will converge towards a single minimum of the cost function. For classifiers in the LVQ-family, this minimum corresponds to a specific subspace of the original feature space, while the remaining subspace may still contain class-relevant information. In this way, in a traditionally trained model, often only a part of the potentially useful class-specific information is used.

In this work, we extend our contribution to the 2023 European Symposium on Artificial Neural Networks, Computational Intelligence, and Machine Learning (ESANN) \cite{lovdal2023improved}. 
\rev{There, 
we presented an extension of Generalized Matrix LVQ (GMLVQ) \cite{schneider2009adaptive, bunte2012limited} and showed} that the successive removal of 
dominantly relevant directions
in feature space and subsequent re-training of GMLVQ with the remaining information allows to infer the most class-relevant subspace. This 
\textit{Iterated Relevance Matrix Analysis} (IRMA)  
facilitates the detailed analysis of feature relevances - especially in presence of multiple weakly relevant features.
Moreover, we demonstrated that the discriminative low-dimensional representation and
visualization of 
labeled data sets could be enhanced compared with the basic GMLVQ approach \cite{schneider2009adaptive, bunte2012limited}. In this work, we extend the feature relevance analysis and discriminative visualization from a binary to a multi-class setting. Furthermore, we investigate the potential of IRMA-based dimensionality reduction, by comparing the performance of a simple \rev{GLVQ classifier \cite{sato1995generalized}} in three different spaces: using no dimensionality reduction, GMLVQ-based and IRMA-based dimensionality reduction. This work being an extension of our conference contribution, some sections have been adopted from the original work without explicit further indication.

\rev{Learning in mutually orthogonal subspaces, similar to the basic idea of IRMA, has 
been considered earlier for Support Vector 
Machines and  Linear Discriminant Analysis,
see e.g.\ 
\cite{tao2008recursive,recursiveLDA},
with emphasis on the dimensionality
reduction as an alternative to 
Principal Component Analysis. This work is also partially building on van Veen et al. \cite{van2024subspace}, where a GMVLQ-based orthogonal direction is learned and projected out to reduce source-specific bias in data.
Here, we focus on exploiting orthogonal
discriminative subspaces for the improved
interpretation of feature relevances and
the potential construction of more robust classifiers. 

Other approaches to the analysis of feature relevances in linear mappings and prototype-based classifiers have been addressed previously
in several studies, see e.g.\ 
\cite{gopfert2017feature,gopfert2018interpretation,Schulz2015,Frenay2014}. 
}

Our paper is structured as follows. In Section~\ref{methods} we introduce the suggested procedure (IRMA), and describe our experimental setup. The illustrative application of IRMA to our artificial and benchmark data sets (binary and multi-class) is presented and discussed in Section~\ref{results}. With this work, we aim to answer the following research questions. First, can IRMA be used to improve the interpretation of feature relevances in classification problems? Additionally, we wish to investigate whether the class-discriminative subspace created by IRMA is able to capture additional relevant information, in order to potentially improve the performance of a classifier. To this end, we compare the performance of a simple GLVQ classifier, and evaluate the second research question: Is IRMA-based dimensionality reduction better than GMLVQ-based? Finally, in Section~\ref{conclusion}, we discuss the potential further ways to exploit the class-specific information extracted by IRMA, and suggested future work.

\section{Methods}\label{methods}
\subsection{Iterated  Relevance Matrix Learning} \label{irma}

An LVQ system assigns $N$-dim.\ feature vectors $\mathbf{x} \in 
\mathbb{R}^N$ to one of $C$ classes labeled by $S \in \{1,2,\ldots,C\}$.
The \textit{nearest prototype} classification is based on the distances 
of $\mathbf{x}$ from a set of $M$ prototypes 
$\{ \mathbf{w}_j \in \mathbb{R}^N\}_{j=1}^M.$ 
Each prototype 
represents one of $C$ classes as denoted by the 
labels $S(\mathbf{w}_j) \in \{1,2,\ldots,C\}$.

GMLVQ in its basic variant \cite{schneider2009adaptive} employs a
global distance measure 
of the form 
\begin{equation}\label{eq:distance} 
    d(\mathbf{w}_j, \mathbf{x}) = (\mathbf{x} - 
    \mathbf{w}_j)^\top \Lambda (\mathbf{x} - \mathbf{w}_j),
    \mbox{~~~with~} \Lambda = \Omega^\top \Omega.
\end{equation}
Here, the relevance matrix $\Lambda\in \mathbb{R}^{N\times N}$ is
re-parameterized in terms of an auxiliary matrix $\Omega\in \mathbb{R}^{N\times N}$
as to guarantee that $\Lambda$ is symmetric and
positive semi-definite  with $d(\mathbf{w}_j,\mathbf{x})\geq 0.$ 
Extensions to local relevance matrices or
rectangular $\Omega$ have been considered in the
literature \cite{schneider2009adaptive,bunte2012limited}. 

Given a set of data $\left\{\mathbf{x}^\mu,S^\mu\right\}_{\mu=1}^P,$
 prototypes $\mathbf{w}_j$ and matrix $\Omega$ are optimized in a training process 
which is guided
by the minimization \rev{of the cost function \cite{sato1995generalized}}
\begin{equation}\label{eq:objective_function} 
    E  = \sum^{P}_{\mu=1}  \phi \left[
    \frac{d^{\Lambda}(\mathbf{w}_+,\mathbf{x}^\mu) - d^{\Lambda}(\mathbf{w}_-,\mathbf{x{^\mu})}}
    {d^{\Lambda}(\mathbf{w}_+,\mathbf{x}^\mu) + d^{\Lambda}(\mathbf{w}_-,\mathbf{x{^\mu})}} \right], 
    \mbox{~~with~}
    \phi(z)=z \mbox{~in the following.}
\end{equation}
For a given example $\left\{\mathbf{x}^\mu,S^\mu\right\}$, 
$\mathbf{w}_+$ denotes the \textit{closest correct} prototype with $d(\mathbf{w}_+,\mathbf{x}^\mu) \leq d(\mathbf{w}_j, \mathbf{x}^\mu)$ 
among all $\mathbf{w}_j$ with 
$S(\mathbf{w}_j)=S^\mu.$ Correspondingly, $\mathbf{w}_-$ is the \textit{closest wrong} prototype carrying a label 
different from $S^\mu.$ In practice, GMLVQ ensures that 
the data points are linearly mapped by $\Omega$ into a 
space where classes are separated as well as possible. 
An additional normalization of
\rev{the form
\begin{equation} 
\sum_{i=1}^N \Lambda_{ii}
= \sum_{i,j=1}^N \Omega_{ij}^\top \Omega_{ji} =1
\label{eq:normalization}
\end{equation}
is imposed in order to avoid}
numerical instabilities and support
comparability \rev{of relevance matrices} \cite{schneider2009adaptive}. 
The resulting diagonal entries $\Lambda_{jj}$ 
quantify the relevance of dimension $j,$ 
provided all features $x_j$ are of the same magnitude \cite{schneider2009adaptive}. 
Throughout the following we achieve this by applying a \rev{feature-wise $z$-score 
transformation in all considered data sets}.

The symmetric semi-definite relevance matrix can be expressed as:
\begin{equation} 
\Lambda=\sum_{j=1}^N \, \lambda_j \mathbf{v}_j \mathbf{v}_j^\top  
\mbox{~with~}
\Lambda \, \mathbf{v}_j = 
\lambda_j \mathbf{v}_j.
\end{equation} 
\rev{The matrix  $\Omega= \sum_{j=1}^N \sqrt{\lambda_j} \mathbf{v}_j \mathbf{v}_j^\top
$
serves as a canonical, symmetric reparameterization of $\Lambda$ in the following. Furthermore, we
 assume that eigenvalues can be ordered as 
$\lambda_1 \geq \lambda_2 \ldots \geq \lambda_N$ without loss of generality. }

After training, the relevance matrix typically assumes a low rank and
is dominated by a few leading eigenvectors, see \cite{biehl2015stationarity} for 
a detailed discussion and analysis. This property facilitates e.g.\ 
the discriminative visualization of the data set in terms of projections
onto the first eigenvectors \cite{schneider2009adaptive, bunte2012limited}. 

In two-class problems, for instance,
the training typically identifies 
a single, most discriminative direction
$\mathbf{v}_1^{(0)}$ with $\lambda_1^{(0)}\approx 1$
and $\Lambda^{(0)} \approx \mathbf{v}_1^{(0)}
\mathbf{v}_1^{(0)\top}.$ Here and in the following the superscript $(0)$ refers
to the results of a first, unrestricted GMLVQ training. 
In such a situation, the eigenvectors $\mathbf{v}_j^{(0)}$ with $j\geq 2$
form an arbitrary basis of the space orthogonal to $\mathbf{v}_1^{(0)}$
with no particular order, and at the end of training this subspace is ignored when the model computes its distances. Note, however, that the
corresponding 
$(N\!-\!1)$-dim.\  subspace 
very likely still contains
relevant information about the classes,
reflecting the potential ambiguity of the relevance assignment. The selection of a particular $\mathbf{v}_1^{(0)}$
may depend strongly on \rev{initial conditions
and on properties of} the actual training data
set, possibly leading to an overfitted relevance analysis. 

In order to obtain more comprehensive insights, 
we can perform a  
second GMLVQ training process which is restricted to an 
orthogonal
subspace by considering a distance measure 
of the form 
(\ref{eq:distance}) with 
$ \Lambda^{(1)}= \Omega^{(1)\top} \Omega^{(1)}$ 
under the constraint
that $\Omega^{(1)} \mathbf{v}_1^{(0)} = 0.$  This can be achieved by
applying the projection  
\begin{equation} \label{eq:correction1}
\Omega^{(1)} \to \Omega^{(1)} \, [{I}-\mathbf{v}_1^{(0)}
\mathbf{v}_1^{(0)\top}]
\end{equation}
after each update step, followed by the 
normalization of $\Omega^{(1)}$ (cf.\ Eq.\ 
\ref{eq:normalization}). In other words, this projection ensures that contributions corresponding to 
$\mathbf{v}_1^{(0)}$ are disregarded in the feature space.
Now, the leading eigenvector $\mathbf{v}_1^{(1)}$
of the resulting $\Lambda^{(1)}$
represents the most discriminative
direction orthogonal to $\mathbf{v}_1^{(0)}.$ 
The degree to which $\mathbf{v}_1^{(1)}$ carries
class relevant information can be evaluated in
terms of a performance measure 
of the restricted classifier, e.g. by the 
balanced accuracy $BAC^{(1)},$ 
estimated in an appropriate validation procedure. \\
Obviously, we can apply the idea iteratively 
and obtain a sequence of vectors $\mathbf{v}_1^{(j)}$ 
each of which is orthogonal to all 
 $\mathbf{v}_1^{(i)}$
with $i=0,1,\ldots, j-1.$ 
In each step $j\geq 1$ of this 
\textit{Iterated Relevance Matrix Analysis} (IRMA)
 we perform
GMLVQ training where the projection 
\begin{equation} 
\label{keystep}
\Omega^{(j)} \to \Omega^{(j)} \left[I - \sum_{i=0}^{j-1} \, \mathbf{v}_1^{(i)}
\mathbf{v}_1^{(i)\top} \right]
\end{equation} 
is applied after each update together with the appropriate normalization. 
We  will refer to the unrestricted GMLVQ training as the \textit{$0$-th iteration}.
The key step (\ref{keystep}) is reminiscent of the subspace
correction in \cite{van2024subspace}, where it however 
serves a different purpose. 

The procedure can be terminated when the classifier in iteration $(k+1)$ achieves
only random or near random classification performance as signaled by, for
example, a $BAC^{(k+1)} \approx 0.5.$  in a binary problem. 
The obtained subspace  
\begin{equation} \label{subspace} 
V= \mathrm{span} \{\mathbf{v}_1^{(0)},\mathbf{v}_1^{(1)}, \ldots \mathbf{v}_1^{(k)}\}
  \mbox{~~with associated projections~} y_i^\mu =\mathbf{x}^\mu \cdot \mathbf{v}_1^{(i)}
\end{equation} 
can be interpreted as to contain (approximately) all
class relevant information 
in feature space. Hence, it can serve for further analysis of feature relevances. 
An obvious application could be the low-dim.\ representation of labeled
data sets in terms of the  
$y_i^\mu,$ 
e.g.\ for the purpose of two- or three-dim.\ visualizations.

\rev{
We would like to stress again that $V$ in Eq.\ (\ref{subspace})
differs significantly from the set of 
leading eigenvectors 
$\{\mathbf{v}_1^{(0)},\mathbf{v}_2^{(0)}, \ldots \mathbf{v}_k^{(0)} \}$ 
as obtained in a single application of unrestricted GMLVQ. There, no particular
order is imposed on the orthogonal vectors $\mathbf{v}_j^{(0)}$ for $j\geq 2.$
In a typical two-class problem only the discriminative power of 
$\mathbf{v}_1^{(0)}$ is represented explicitly. }

When applying IRMA in a multi-class setting, multiple relevant eigenvectors per iteration could be removed. For multi-class problems, the converged 
$\Lambda$ \rev{is typically dominated by a set of (several) relevant eigenvectors \cite{schneider2009adaptive}. Their 
number} is dependent on the number of classes and the properties of the data. The eigenvalue profile of $\Lambda$ can be inspected in order to make a decision regarding \rev{the number $K$ of eigenvectors that should be removed in each
iteration by applying
\begin{equation} \label{eq:multipleremoval}
\Omega^{(j)} \to \Omega^{(j)} \left[I - \sum_{i=0}^{j-1} \, \sum_{l=1}^K \mathbf{v}_l^{(i)}
\mathbf{v}_l^{(i)\top} \right].
\end{equation}
Here, $\mathbf{v}_l^{(i)}$ denotes the
$l$-th leading eigenvector of the
relevance matrix obtained in IRMA iteration $i.$ }

\rev{The choice of $K$ also depends
on the actual motiviation for applying IRMA. 
If the goal is a thorough analysis of feature
relevances, it is favorable to inspect models
operating in mutually orthogonal subspaces. 
Alternatively, by removing single eigenvectors
in each iteration, the classifiers will use
partially overlapping information, with possibly 
better performance but harder to interpret relevances. 
To inspect models operating in orthogonal subspaces, one could remove a
variable number of eigenvectors such that
the sum of their eigenvalues is close to 1. On the other hand, if the goal is to construct a class-specific subspace, one or a fixed number of
multiple eigenvalues can be removed per iteration. Removing one eigenvector per iteration will be slightly less efficient than removing multiple at the time, while it potentially will be more precise with respect to
maximizing the class separation in the resulting IRMA-subspace.
}

\subsection{Experiments}

We will use three different illustrative data sets to demonstrate the properties of IRMA: One artificial data set drawn from \rev{a mixture of two Gaussians}, and two data sets from the UCI machine learning repository (one two-class and one seven-class). For simplicity and increased interpretability of feature relevance profiles, we \rev{apply a $z$-score transformation to all features} once before the start of training.

For all three data sets, we estimate the BAC per iteration (or number of orthogonal solutions with a reasonable performance) by performing a 30 times repeated application of IRMA with a 50/50 train-test split. The training and test sets are determined by stratified random sampling for each experiment.

We will inspect both the discriminative visualizations for each data set, projecting the data onto the eigenvectors obtained by GMLVQ or \rev{IRMA. For this purpose, we display the result of one arbitrary training process in the validation scheme. In addition, we inspect the feature} relevance profiles, \rev{as given by the diagonal elements of $\Lambda$}, per iteration for the real-world data sets. 
\rev{To this end,} we apply IRMA once on the full set of available data. For binary problems, we remove one eigenvector per iteration, and for the multi-class problem, we remove multiple per iteration\rev{, such that the summed eigenvalues indicate that most of the subspace relevant for the solution in question is covered}.

Additionally, we investigate the suitability of IRMA for dimensionality reduction, using the two real world data sets. For this purpose, we compare the performance of a simple GLVQ classifier in (a) original data space, (b) GMLVQ-space, and (c) IRMA-space. As $\Lambda$ typically converges to a low-rank representation, it is known that traditional GMLVQ can also be used as a form of dimensionality reduction, by projecting the data onto the leading eigenvectors of $\Lambda$. We are interested in whether the iterative approach by IRMA may capture more relevant information than a \rev{single 
solution identified by GMLVQ}. Therefore, we again use a 50/50 train-test split, and measure the performance in terms of  balanced accuracy. For (a), we apply GLVQ without any dimensionality reduction. For (b), we apply GMLVQ on the training set, project all data onto the eigenvectors comprising 99\% of the summed eigenvalues, and then train a GLVQ classifier in the lower-dimensional space using the same set of training data. In (c), we apply IRMA on the training set, and project all data into the growing subspace 
\rev{combined from the iterations of the method.} We then retrain the GLVQ classifier in \rev{this} IRMA-space, using the same training set. For extracting this "class-specific subspace", we remove one eigenvector per iteration for both the two-class and seven-class data set. We perform experiments a-c for 1, 2 and 3 prototypes per class, in order to observe how the performance changes with increasing flexibility of the GLVQ model. For each training and test round, the GLVQ, GMLVQ and IRMA models are assigned the same number of prototypes. We specifically employ a simple GLVQ model to test our dimensionality reduction hypothesis, as reapplying GMLVQ in IRMA-space would most likely converge back to the initial solution of the 0-th iteration.

We use the same standard parameters for all LVQ-based models in the experiments: \rev{$30$ epochs of stochastic gradient descent, activation function identity, and initial step sizes} of $0.1$ and $0.01$ for the prototypes and relevance matrix, respectively. The rest of the parameters are left as the default values, as implemented by the Python sklvq package \cite{van2021sklvq}.

Below, we cover further details of the data sets, as well as the results and discussion of the experiments.


\section{Results and Discussion}\label{results}


\begin{figure}[h!]
\centering
\includegraphics[trim = 0 0 0 0.55cm, clip, width=0.3\textwidth]{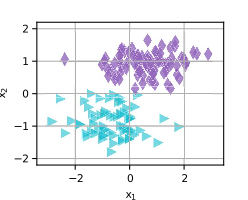}
\put(-95,-6){(a)}
\mbox{~~}
\includegraphics[width=0.3\textwidth]{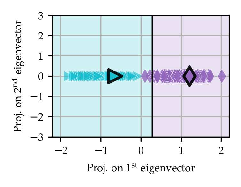}
\put(-95,-6){(b)}
\mbox{~~}
\includegraphics[width=0.3\textwidth]{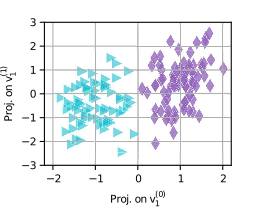}
\put(-95,-6){(c)}
\caption{Artificial data: original features $x_1, x_2$ of the 
data set (a), projections on $v_1^{(0)}$,
$v_2^{(0)}$ of unrestricted GMLVQ 
(b), and projections on the eigenvectors 
$v_1^{(0)}$ and 
$v_1^{(1)}$
of the unrestricted system and the first iteration of IRMA 
in (c).} \label{potatoes}
\end{figure}

\subsection{Artificial Data}
We first consider an extremely simple and clear-cut artificial
two-class data set illustrated in Fig.\ \ref{potatoes} (a). 
Feature vectors $\mathbf{x}\in\mathbb{R}^{4}$ comprise
two informative components $x_1, x_2$ in which 
each class corresponds to an elongated Gaussian cluster \rev{with means $\mu_1 = [-1, -8],\ \mu_2 = [1, 8]$ and covariance matrix $\Sigma = \big(\begin{smallmatrix}
  2 & 0\\
  0 & 12
\end{smallmatrix}\big)$ for both clusters}. 
The remaining components are independently 
drawn from an isotropic
zero mean, unit variance normal density, before applying the $z$-score transformation. 
As can be seen in panel (a), feature $x_2$ should be sufficient 
to  separate the classes with almost 100\% accuracy. However, 
classes also separate along $x_1$, albeit less perfectly. 
Unrestricted GMLVQ with
one prototype per class realizes near
perfect classification with $BAC^{(0)} \approx 0.99$
(w.r.t.\ training and test) in a balanced data set 
of 600 samples, where the training 
set contains $300$ randomly drawn examples and 
the remaining $300$ form a 
test set. Projections on the
leading eigenvectors are shown in panel (b) of Fig.~\ref{potatoes}. 
The dominating eigenvector is
$\mathbf{v}_1^{(0)} \approx (0.18,0.98, -0.02, 0.01)^\top$ corresponding to $\Lambda^{(0)}_{jj}
\approx \delta_{j,2}$. The orthogonal 
$\mbox{v}_2^{(0)}$ is essentially random as 
indicated by the absence of a separation of classes, resulting 
in an effectively one-dim.\ visualization. 

In the first IRMA iteration, the leading eigenvector
of $\Lambda^{(1)}$ approaches the second relevant direction:
$\mathbf{v}_1^{(1)} \approx (0.98,-0.18, -0.02, -0.02)^\top$
with $\Lambda_{jj}^{(1)}\approx \delta_{j,1}.$ 
As expected, the performance drops compared to the unrestricted
system: we observe a $BAC^{(1)}$ of
$0.70$ (training) and $0.68$ (test).
As shown in panel (c) of Fig.\ \ref{potatoes}, the projections 
$y_0,y_1$, cf.\ Eq.\  (\ref{subspace}), of
the data set onto $\mathbf{v}_1^{(0)}$ and 
$\mathbf{v}_1^{(1)}$ display both
relevant separating directions and reproduce the 
cluster structure
of the original features $x_1, x_2.$
Already in the second iteration of IRMA, the 
accuracy drops to $BAC^{(2)} \approx 0.52$ and $0.51$ for training and 
test data, respectively. As expected,
no further relevant directions can be identified.

\subsection{Wisconsin Diagnostic Breast Cancer data}
\begin{figure}[h!]
\centering
\includegraphics[width=0.29\textwidth]{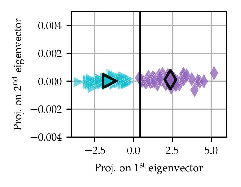}
\put(-95,-6){(a)}
\mbox{~~}
\includegraphics[width=0.29\textwidth]{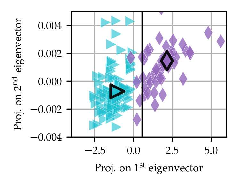}
\put(-95,-6){(b)}
\mbox{~~}
\includegraphics[width=0.29\textwidth]{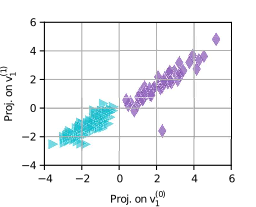}
\put(-95,-6){(c)}
\caption{Wisconsin data set: Projections after 0th (a), 1st iteration (b), and data projected onto leading eigenvectors of 0th and 1st iteration, respectively (c).} \label{fig:wisconsin_projections}
\end{figure}

This benchmark data set from the
UCI Machine Learning Repository \cite{lichman2013uci, misc_breast_cancer_wisconsin_(diagnostic)_17} 
contains $569$ samples with $30$ features extracted from cells in an image of a fine needle aspirate of a breast mass  (357 benign, 212 malignant). Fig.~\ref{fig:wisconsin_projections} shows the projection of part of the training data into GMLVQ space at the end of training for the unrestricted system (iteration 0, (a)), and after the 1st iteration (b). Here, the benign samples are displayed as cyan triangles, and the malignant as purple diamonds. Fig.~\ref{fig:wisconsin_projections} (c) shows the training data projected onto the leading eigenvector of the 0th and 1st iteration, where you can see a clear discrimination of the two classes along both coordinate axes. The leading eigenvalue of the GMLVQ system from both the 0th and 1st iteration is $ \approx 1.0$ respectively, meaning that there is no contribution from non-dominant eigenvectors from iteration 0 in iteration 1.

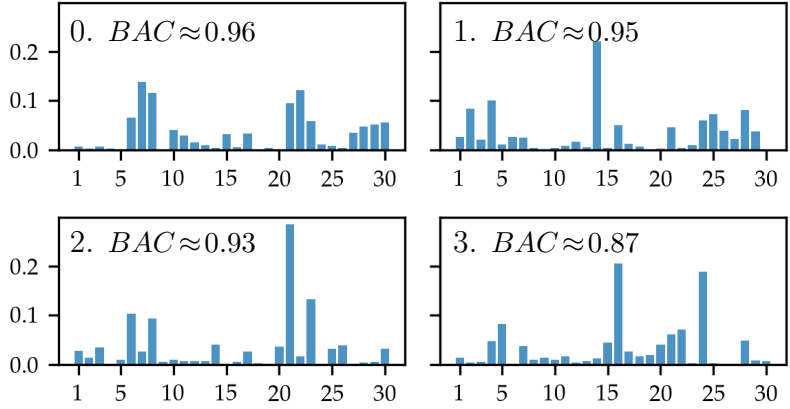
\begin{figure}[t!]
\centering
\includegraphics[width=0.8\textwidth]{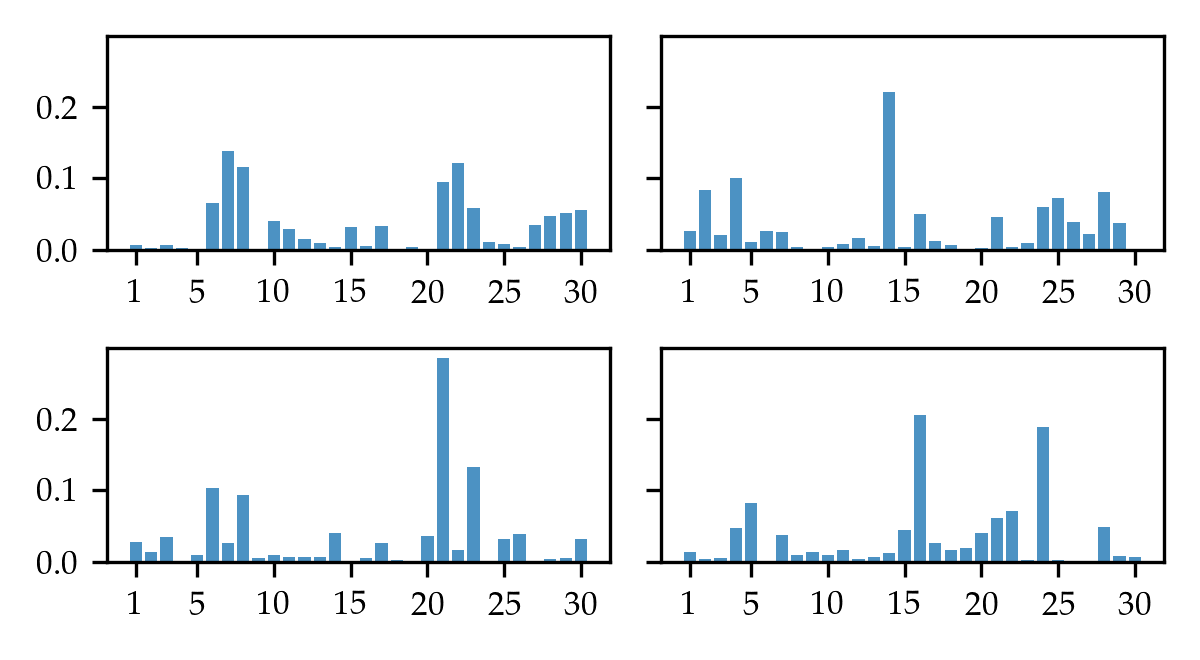}
\put(-280,148){0. \small $BAC\!\approx\!0.96$}
\put(-135,148){1. \small $BAC\!\approx\!0.95$}
\put(-280,68){2. \small $BAC\!\approx\!0.93$}
\put(-135,68){3. \small $BAC\!\approx\!0.87$}
\caption{Wisconsin data set: Diagonal of $\Lambda$ per iteration $(i)$, which is indicated as $i$ in the \rev{upper left} corner of 
each panel.
In addition, the obtained random sampling validation $BAC$ w.r.t.\ test data are shown.} \label{fig:wisconsin_lambda}
\end{figure}

The application of IRMA allows deeper insights
into the feature relevances. For example, Fig.\ \ref{fig:wisconsin_lambda} shows
that features 4 and 14 display significant $\Lambda_{jj}>0.1$ 
in iteration $(1)$ (being the most important features), while 
they appear 
irrelevant in the unrestricted system $(0).$
However, the performance of the two systems
is virtually identical 
with $BAC^{(1)} \approx BAC^{(0)}$.
Hence, these features constitute examples of 
\textit{weakly} relevant dimensions in the sense of the discussion
given in \cite{gopfert2017feature, gopfert2018interpretation}: they enable successful classification
in $(1)$, but are replaced by other (combinations of)
features  
in $(0)$. 
Similarly, 
the single feature $j=21$ dominates the 
classification in iteration (2), while it plays only a 
minor role in the other classifiers.

Note that the test set accuracies
decrease to 
$BAC^{(4)}\approx 0.84 $
and $BAC^{(5)}\approx 0.78 $, $BAC^{(6)}\approx 0.74 $, $BAC^{(7)}\approx 0.70 $, $BAC^{(14)}\approx 0.56$.
Here, we restrict the discussion to $V=\{\mathbf{v}_1^{(0)},\mathbf{v}_1^{(1)},
\mathbf{v}_1^{(2)}, \mathbf{v}_1^{(3)}\}$ as the
most discriminative subspace. Five features ($j=9, 12, 13, 18, 19$) display diagonal relevances
$\Lambda_{jj}^{(i)}<0.02$ for all $i\leq 3$ 
and, therefore, could be considered irrelevant. 
Two features ($j=7, 21$) were rated relevant with
$\Lambda_{jj}^{(i)}\geq 0.02$ for all $i\leq 3$. \\

Table~\ref{tab:GLVQ_performance} presents the balanced accuracy scores when training a GLVQ classifier in original data space, GMLVQ-space, and IRMA-space, for 1, 2 and 3 prototypes. GLVQ trained in the original data space performs consistently worse than the GLVQ models trained in one of the lower-dimensional data spaces (BAC $\approx 0.90-0.92$ vs. $0.96-0.97$). GLVQ trained in IRMA space is marginally better than GLVQ trained in a data space derived from GMLVQ, and the scores improve slightly with a larger number of prototypes. The highest score was obtained by training GLVQ with three prototypes in two-dimensional IRMA space (BAC $\approx 0.97$). Interestingly, GLVQ in two-dimensional IRMA space with three prototypes performs marginally better than classical GMLVQ with three prototypes (the latter obtaining an average BAC of 0.963), despite GMLVQ featuring the added possibility of weighing the coordinate axes.

\begin{table}[t!]
\begin{tabular}{lllll}
\toprule
\textbf{Data set}     & \textbf{n$_{p}$} & \textbf{Original} & \textbf{GMLVQ space}   & \textbf{IRMA space}             \\
\midrule
Wisconsin    & 1             & 0.900 (0.02)          &  0.956 (0.02) & 0.958 (0.01), 1-dim           \\
             & 2             & 0.913 (0.02)          &  0.963 (0.01)  &  0.964 (0.01), 2-dim          \\
             & 3             & 0.921 (0.02)          &  0.962 (0.01)    &  \textbf{0.965} (0.01), 2-dim         \\
\midrule
Segmentation & 1             & 0.856 (0.01)          & 0.877 (0.01) & 0.870 (0.01), 7-dim\\
             & 2             & 0.871 (0.01)          & 0.879 (0.02) & 0.889 (0.01), 7-dim\\
             & 3             & 0.878 (0.01)          & 0.894 (0.02) & \textbf{0.898} (0.01), 6-dim \\
\bottomrule
\end{tabular}
\caption{Comparison of GLVQ performance when varying the dimensionality reduction technique and number of prototypes (n$_p$). The balanced accuracy (with standard deviation within brackets) is reported for a 30-times repeated random sampling validation where the classifier has been trained in three different spaces: Original data space (using no dimensionality reduction), GMLVQ-space (using GMLVQ-based dimensionality reduction), and IRMA-space (using IRMA-based dimensionality reduction).}\label{tab:GLVQ_performance}
\end{table}

\subsection{Segmentation data}

The segmentation data set from the UCI machine learning repository \cite{lichman2013uci, misc_image_segmentation_50} is based on a set of seven outdoor images. Features related to color, contrast, hue, saturation, location in the image, and line segments were extracted from $3\times$3 pixel regions. Each sample is labelled as one of seven classes: brickface, sky, foliage, cement, window, path or grass. We merged the original division of training and test set of the repository into a single data set, due to the original split having roughly a 10/90 proportion. We excluded a feature describing the number of pixels per region, as this had the same value for every sample, resulting in 18 remaining features.

\begin{figure}[h!]
\centering
\includegraphics[width = 0.8 \textwidth]{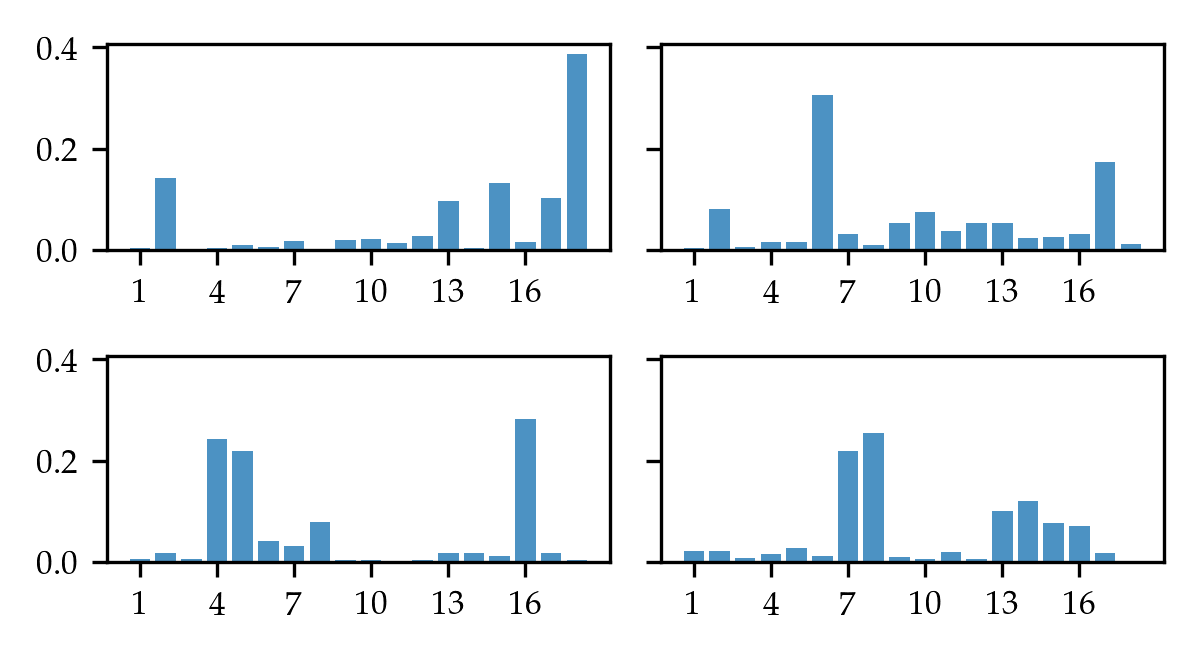}
\put(-280,147){\small 0. $BAC\!\approx\!0.89$}
\put(-83,147){\small 1. $BAC\!\approx\!0.81$}
\put(-280,66){\small 2. $BAC\!\approx\!0.66$}
\put(-83,66){\small 3. $BAC\!\approx\!0.52$}
\caption{Segmentation data set: Diagonal of $\Lambda$ per iteration $(i)$, which is indicated as $i$ 
\rev{in 
each panel}, where IRMA has been applied using all available data. Three eigenvectors are removed per iteration for this seven-class problem, and the average $BAC$ w.r.t.\ test data is indicated on top of each panel.} \label{fig:segmentation_lambda}
\end{figure}

For feature relevance analysis of the segmentation data set, we remove three eigenvectors per \rev{IRMA} iteration. Averaging 10 times repeated experiments, this covers 89\% of the summed eigenvalues for iteration 0, and 97\% in iterations 1-3. Even though 
\rev{one might consider the removal of} four or five eigenvectors after iteration 0, 
\rev{with eigenvalues summing up to 0.96 or 1.0,} 
respectively, we \rev{select three eigenvectors consistently for illustrative purposes.} 
 We display the relevance profiles of the first four iterations in Figure~\ref{fig:segmentation_lambda}. The six iterations of IRMA obtain corresponding average BACs of 0.89 0.81, 0.66, 0.52, 0.30 and 0.16. Note that it is not possible to run more iterations of IRMA on this 18-dimensional data set, as we would remove $18$ \rev{eigenvectors with all data projected onto the origin in  iteration (7)}. As the performance drops considerably from iteration $(1)$ to $(2)$, we could consider the first two iterations 
\rev{and the corresponding six-dimensional subspace} to carry the majority of class-specific information. In the corresponding relevance profiles in Figure~\ref{fig:segmentation_lambda}, we see a similarly interesting pattern as for the Wisconsin data set, where e.g. feature six seems irrelevant in iteration 0, while it is by far the most important feature in iteration 1. Features $j=1$ and $j=3$ are irrelevant for all four iterations displayed, with $\Lambda_{j,j} \leq 0.02$. 

\begin{figure}[t!]
\centering
\includegraphics[width=0.36\textwidth]{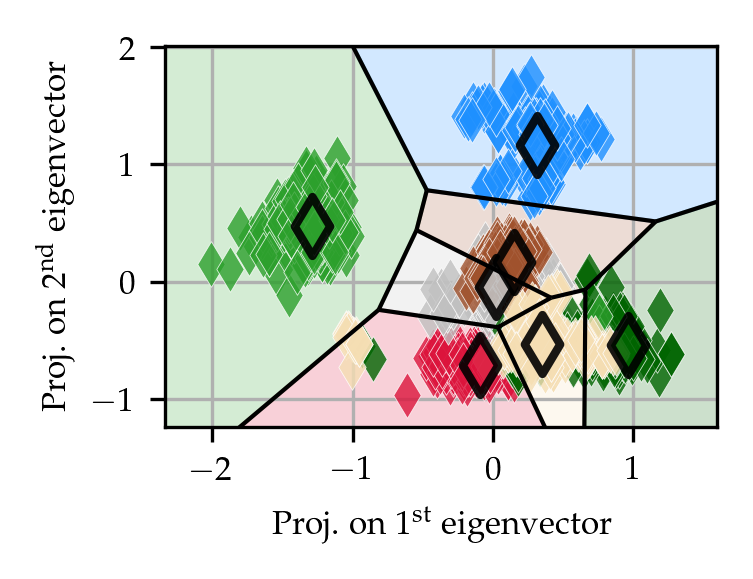}
\put(-60,-10){(a)}
\mbox{~~}
\includegraphics[width=0.29\textwidth, trim = 1cm 0 0 0, clip]{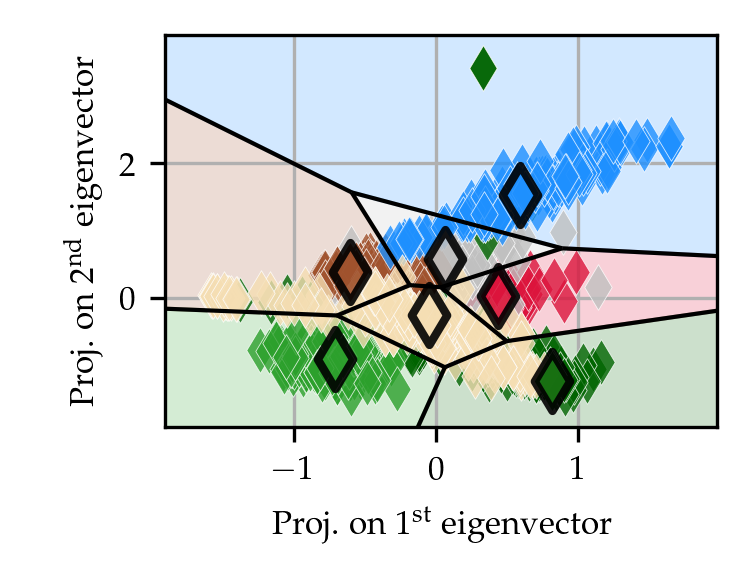}
\put(-60,-10){(b)}
\mbox{~~}
\includegraphics[width=0.29\textwidth, trim = 1cm 0 0 0, clip]{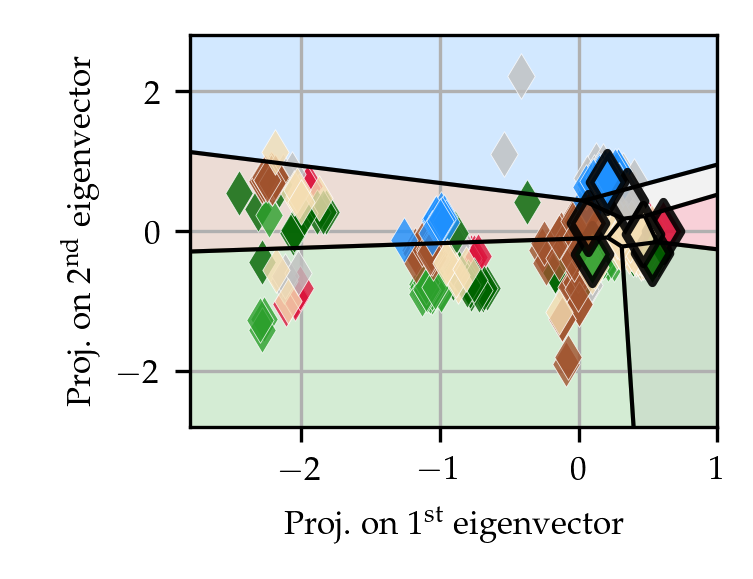}
\put(-60,-10){(c)}
\caption{Segmentation data set: Projections after 0th (a), 1st iteration (b) and 2nd iteration (c). The main cluster was zoomed in on in (c), cutting out a few outliers.} \label{fig:segmentation_projections}
\end{figure}

Fig.~\ref{fig:segmentation_projections} displays the discriminative projections of the three first iterations of IRMA, i.e., the training data projected onto the two leading eigenvectors of each model resulting from an iteration. Fig.~\ref{fig:segmentation_projections} (a) and (b) reveal a good visual separation between the classes for the first two iterations, while the performance deteriorates visibly in (c). For this third iteration, much of the class-relevant information already seems to be absent, reflected by a more chaotic display of the class separation landscape.

\rev{As for the Wisconsin} data set, Table~\ref{tab:GLVQ_performance} displays the average BAC scores obtained when evaluating the suitability of IRMA for dimensionality reduction, by training a simple GLVQ classifier in the original data space, GMLVQ-space, and IRMA-space. Again, the GLVQ classifiers fair better \rev{after  dimensionality reduction by either GMLVQ or IRMA:} $BAC \approx 0.86-0.88$ vs $0.87-0.90$. While GLVQ in GMLVQ-space was slightly better than in IRMA-space when using one prototype, the highest score ($BAC = 0.898$) was obtained in six-dimensional IRMA space using three prototypes per class. Still, this was not better than applying classical GMLVQ with three prototypes per class, obtaining an average BAC of 0.911. Most likely, this is due to the additional flexibility of GMLVQ compared to GLVQ, i.e.\ the possibility of weighting the coordinate axes of the model.




\section{Conclusion and Outlook} \label{conclusion} 

We have shown how IRMA based on GMLVQ with iterative subspace elimination
can be used to find 
class-relevant subspaces for both binary and multi-class classification problems. As an example, we have demonstrated that two mutually exclusive directions provide the same highest performance for the Wisconsin data set. Consequently, feature 
profiles from each relevant subspace can be taken into account for the final feature 
relevance analysis. This should be especially important for data sets with correlated or 
multiple weakly relevant features, or problems where only a small amount of training data is 
available. For the seven-class segmentation data set, we found two distinct subspaces providing a good classification performance. These two subspaces may have had a minor overlap, reflected by that we removed eigenvectors with summed eigenvalues of $\approx 0.89$ after iteration $(0)$.

Additionally, we have demonstrated the potential of using IRMA for dimensionality reduction. Our results show that IRMA-based dimensionality reduction in general may be slightly better than GMLVQ-based, and that it is clearly better than applying GLVQ with no dimensionality reduction at all. Still, the improvements were marginal, and future work may investigate the conditions under which IRMA may provide a clear advantage. It is possible that the data sets included in this work were not sufficiently complex, so that traditional GMLVQ already provided a near optimal solution considering performance. Overall, the application of training a new classifier in IRMA space seems promising: It is able to capture more class-specific information than traditional GMLVQ, enabling slightly enhanced performance even with a simple classifier such as GLVQ. Considering that the classification performance increased when increasing the number of prototypes trained in IRMA-space, we consider it highly interesting future work to evaluate the performance of more complex models in IRMA space. Note that for nonlinear classifiers, iteratively constructing a class-discriminative subspace is nontrivial. While \rev{intrinsically linear models such as GMLVQ and IRMA} may be restricted by their limited complexity, applying more complex models in lower-dimensional IRMA space might offer performance enhancement by allowing flexible decision boundaries to form in a lower-dimensional data space where a maximum amount of class-relevant information is preserved.


Note that at each stage of IRMA a different classifier
is obtained. In particular, the respective 
prototypes are placed in entirely
different positions in feature space. Hence, it is non-trivial
to construct a single classifier from the individual results.
In a binary problem, the naive application of an LVQ classifier on the vectors
$(y_0^\mu,y_1^\mu,\ldots,y_{k}^\mu)^\top$, cf.\ (\ref{subspace}), 
will simply recover the unrestricted classifier by identifying 
$y_0$ as the most discriminative projection. 
Creating a weighted ensemble from all models (iterations) that achieve high performance, may result in a more robust performance and would be of particular interest in the presence of subclusters within the classes. The suitability of IRMA for the purpose of improving performance of classifiers may still be dependent on the geometry of the cost function landscape for a particular data set, \rev{especially if multiple local minima are present. Note that in both real world data sets considered here, the performance of the plain GMLVQ classifier is very good or even near optimal already. In future studies we will aim at more difficult classification problems, in order to fully explore the potential improvement 
by IRMA-based classifiers.}

Furthermore, the issue of creating \rev{a weighted} accumulated relevance profile reflecting the 
importance of a feature across all relevant subspaces is also nontrivial, since not all iterations have the same discriminative accuracy. Note that the results of previous feature relevance analyses depend strongly on the details of the method
and the considered classifiers, compare e.g. \cite{gopfert2017feature} and \cite{gopfert2018interpretation}. We leave the creation of an accumulated relevance profile, \rev{as well as the formal evaluation of stopping criteria for the IRMA iteration, as future work.}

 


\section{Acknowledgements}
SL acknowledges support by the Dutch Stichting ParkinsonFonds (grant number 2022/1891).





 \bibliographystyle{elsarticle-num} 
 \bibliography{cas-refs}





\end{document}